\pdfoutput=1
\documentclass[]{collinear}

\author{Muyu He}
\author{Muhammad Ali Shafique}
\author{Anand Kumar}
\author{Tsach Mackey}
\author{Nazneen Rajani}

\abstract{
Distilling the thinking traces of a Large Language Model (LLM) with reasoning capabilities into a smaller model has been proven effective. 
Yet, there is a scarcity of work done on how model performances scale with the quantity of distillation data.
In this work, we study the scaling trend of distilling competitive coding skills on two small non-reasoning LLMs.
We validate the hypothesis that there is a \emph{valley of code reasoning}:
downstream performance on competitive coding first drops as data quantity increases, then it steadily increases in a sharper-than-log-linear fashion.
Having identified the trend, we further fine-tune the models at two different distillation stages on the same data to ground conclusions on their respective learning phases.
We learn that across stages in the low and medium-low data regimes, small models benefit significantly from easier coding questions than from harder ones.
We also find that, surprisingly, the correctness of outputs in training data makes no difference to distillation outcomes.
Our work represents a step forward in understanding the training dynamics of code reasoning distillation outside intuition.
We are open-sourcing dataset splits used for all our experiments at \url{https://collinear.ai/valley-of-reasoning-data} .}
\correspondence{\url{research@collinear.ai}}
\date{\today}

\usepackage[utf8]{inputenc} % allow utf-8 input
\usepackage[T1]{fontenc}    % use 8-bit T1 fonts
\usepackage{hyperref}       % hyperlinks
\usepackage{url}            % simple URL typesetting
\usepackage{booktabs}       % professional-quality tables
\usepackage{amsfonts}       % blackboard math symbols
\usepackage{nicefrac}       % compact symbols for 1/2, etc.
\usepackage{microtype}      % microtypography
\usepackage{xcolor}         % colors
\usepackage{graphicx}
\usepackage{wrapfig,lipsum,booktabs} % put this in your preamble
\usepackage{multirow}
\usepackage{makecell}
\usepackage{fvextra,times}
\usepackage{subcaption}
\usepackage{booktabs}
\usepackage{ulem} % for \uline (underline)
\makeatletter
\def\blfootnote{\xdef\@thefnmark{}\@footnotetext}
\makeatother

\title{The Valley of Code Reasoning: Scaling Knowledge Distillation of Large Language Models}

\begin{document}

\maketitle

% \begin{abstract}
% Distilling the thinking traces of a Large Language Model (LLM) with reasoning capabilities into a smaller model has been proven effective. 
% Yet, there is a scarcity of work done on how model performances scale with the quantity of distillation data.
% In this work, we study the scaling trend of distilling competitive coding skills on two small non-reasoning LLMs.
% We validate the hypothesis that there is a \emph{valley of code reasoning}:
% downstream performance on competitive coding first drops as data quantity increases, then it steadily increases in a log-linear fashion.
% Having identified the trend, we further finetune the models at two different distillation stages on the same data to ground conclusions on their respective learning phase.
% We learn that across stages in the low and medium-low data regimes, small models benefit significantly from easier coding questions than from harder ones.
% We also find that surprisingly, the correctness of outputs in training data makes no difference to distillation outcomes.
% Our work represents a step forward understanding the training dynamics of code reasoning distillation outside intuition.
% \end{abstract}
\blfootnote{Accepted to the NeurIPS 2025 workshop on Deep Learning for Coding (DL4C)}
\section{Introduction}

Test-time scaling techniques have enabled large language models (LLMs) to perform complex multi-step reasoning through chain-of-thought (CoT) style processes. 
Models such as DeepSeek-R1~\cite{guo2025deepseek} and QwQ-32B~\cite{qwq32b} demonstrated strong reasoning abilities, particularly in the domains of mathematics, science and coding. 
These, in turn, have amplified efforts to distill test-time compute gains via supervised fine-tuning (SFT) into smaller fine-tuned models. 
Recent work has demonstrated that LLMs can be taught to produce such long CoT reasoning through distillation in a surprisingly data-efficient manner~\cite{Muennighoff2025s1, ye2025limo}. 
For example,~\cite{li2025llmsReasonFromDemos} fine-tuned a 32B model on only $17k$ reasoning samples and achieved dramatic gains on competitive coding benchmarks comparable to larger models.
Despite the growing body of work on curating high-quality post-training data for reasoning distillation for frontier coding performance ~\cite{guha2025openthoughts,ahmad2025opencodereasoning}, only limited focus has been given to the training dynamics of these student models and how these dynamics influence the acquisition of reasoning abilities.

\begin{figure}[ht] 
\includegraphics[width=\linewidth]{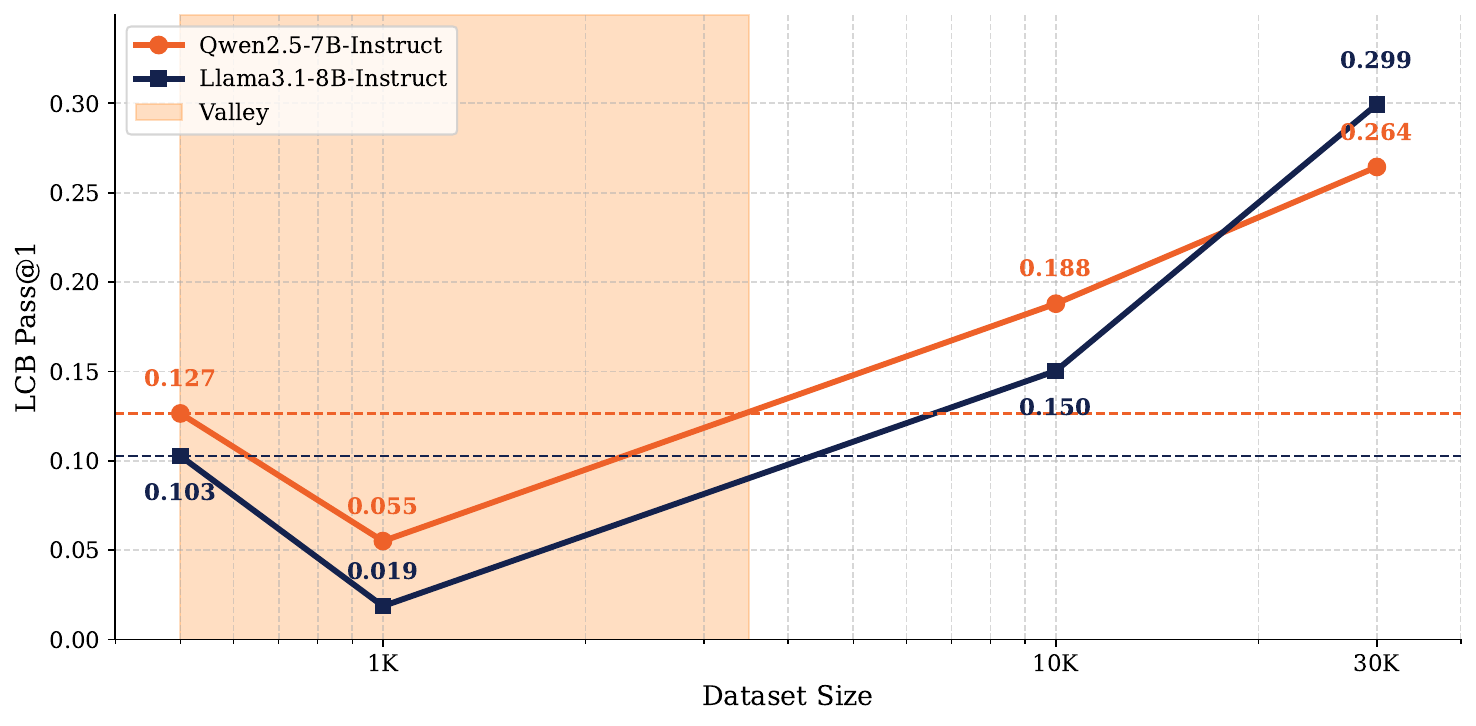} 
\caption{ Evaluation scores on LCB of two distilled small models show that performances initially decreases by half but then steadily increases in a log-linear trend toward the 30K data upper bound.} \label{fig:valley} 
\end{figure}
In this work, we investigate how reasoning abilities are distilled into non-reasoning student models, specifically asking whether their reasoning abilities emerge linearly. 
We hypothesize that, for small non-reasoning LLMs in low to medium-low data regime, distillation does not yield monotonic improvements: 
performance initially declines when trained on small data, then steadily improves as data scales, which we refer to as the \textit{valley of reasoning}. 
Grounding our findings on model performances on \textsc{LiveCodeBench} (LCB) ~\cite{jain2024livecodebench}, a competitive coding benchmark, we validate this claim and observe a sharper-than-log-linear trend once the model gets out of the valley (Fig \ref{fig:valley}). 

Inspired by the findings of ~\cite{li2025llmsReasonFromDemos}, which argues that output structure rather than code correctness in data drives improvement, we further whether this conclusion holds uniformly across stages of distillation. 
We find that both during the non-reasoning phase and the intermediate phase where the model acquires decent reasoning skills, code correctness in data has little impact on evaluation performances.
However, for both stages, we consistently observe that distillation data containing easier coding questions outperform the one with harder questions.
This suggests that at a low data regime, small models can better study after easier examples to yield immediate gains.

% We hypothesize that this early dip reflects a stage in which the model is acquiring the structure of reasoning chains (e.g., reflection, multi-step decomposition) (phase 1 in the Fig.~\ref{fig:valley}). This interpretation is supported not only by lower benchmark accuracy relative to the base model but also by reduced completion rates on coding tasks such as the  LiveCodeBench (LCB)~\cite{jain2024livecodebench}. With further scaling of distillation data, performance and completion rates improve steadily. We propose that in this later stage, the content of training data becomes crucial: continued improvements require reasoning traces that are not only increasingly challenging but also factually correct (phase 2 in the Fig.~\ref{fig:valley}).

% We show that the post-training data quality is dependent on the abilities of the student model, whether the student model has no reasoning, decent reasoning or good reasoning capabilities.

% \begin{itemize}
%     \item No reasoning
%     \item Decent reasoning
%     \item Good reasoning
% \end{itemize}

\section{Related work}

\paragraph{SFT for Reasoning Distillation} SFT on reasoning traces generated by reasoning models has proved effective at improving smaller models~\cite{guo2025deepseek}. Motivated by this, many works have investigated methods for performing distillation. In particular, recent data-efficiency results show that small, carefully curated sets can already induce strong reasoning behavior in large (32B+) models~\cite{ye2025limo,li2025llmsReasonFromDemos}, complementing early self-improvement/distillation antecedents~\cite{zelikman2022star,zelikman2024quietstar}. Several studies probe the components of reasoning traces: prefix-only supervision can enable highly efficient distillation~\cite{ji2025upft}; reasoning quality of the distillation traces can impact downstream performance~\cite{luo2025dlcot,yu2025lsmixture}; and the CoT lengths in the mixture can be tuned to optimize both token efficiency and model capabilities ~\cite{wu2025moreisless,ma2025cotvalve,xu2025chainofdraft}. Strikingly, ~\cite{li2025llmsReasonFromDemos},~\cite{ahmad2025opencodereasoning}, and~\cite{guha2025openthoughts} have found that for models in low-data regimes, the correctness of the reasoning trace does not impact distillation quality. However, these works do not investigate how data quality impacts different stages of training, a matter of practical importance in large-scale post-training processes.

\paragraph{Scaling Distillation for Coding} OpenThoughts~\cite{guha2025openthoughts}, OpenCodeReasoning (OCR) ~\cite{ahmad2025opencodereasoning}, and rStar-Coder~\cite{liu2025rstar} find that competitive coding performance can be improved by scaling the reasoning dataset to millions with high question duplication counts. 
These works evaluate on robust, contamination-controlled code benchmarks such as LCB~\cite{jain2024livecodebench} and CodeElo~\cite{quan2025codeelo}, reinforcing the observed SFT-only gains. Together, they demonstrate the potential of large-scale SFT for dramatic improvements in base-model capabilities, but they do not focus on the model performance during intermediate data quantities. This motivates us to conduct an in-depth study on the training dynamics and progress on downstream evaluations.

\section{The Valley of Reasoning}
\label{sec:valley_of_reasoning}

In regards to the training dynamics of reasoning distillation, we want to answer three research questions (RQs):
(i) Does code reasoning performance scale \emph{monotonously} with data quantity?,
(ii) Does the correctness of the teacher's responses matter for scaling student performance? and
(iii) Does the difficulty of the input problems have an impact on the student performance?
We now discuss how these three questions inform our data curation strategies, and detail the findings in section~\ref{sec:findings}.

\renewcommand{\arraystretch}{1.25}

\begin{table*}[t]
\centering
\small
\setlength{\tabcolsep}{8pt}

% ---------- LEFT TABLE ----------
\begin{minipage}[t]{0.47\textwidth}
% \vspace{0pt}
\centering
\begin{tabular}{llcc}
\toprule
\textbf{Size} & \textbf{Metric} & \textbf{Qwen 2.5} & \textbf{Llama 3.1} \\
\midrule
\multirow{2}{*}{1K}  
 & Completion rate  & 0.1842 & 0.4649 \\
 & Think tag rate & 0.1406 & 0.0777 \\
\midrule
\multirow{2}{*}{10K} 
 & Completion rate  & 0.3509 & 0.3459 \\
 & Think tag rate & 0.3496 & 0.3496 \\
\midrule
\multirow{2}{*}{30K} 
 & Completion rate  & 0.5802 & 0.7080 \\
 & Think tag rate & 0.5677 & 0.7068 \\
\bottomrule
\end{tabular}
\caption{
Completion and think tag rates of Qwen2.5 and Llama3.1 across training data sizes.
}
\label{tab:qwen_llama_nested}
\end{minipage}
\hfill
% ---------- RIGHT TABLE ----------
\begin{minipage}[t]{0.47\textwidth}
% \vspace{0pt}
\centering
\begin{tabular}{lcc}
\toprule
\textbf{Dataset} & \textbf{Qwen 2.5} & \textbf{Qwen 2.5-30K} \\
\midrule
Baseline     & 0.126 & 0.264 \\
\midrule
Correct 6K   & \uline{0.185} & 0.347 \\
Incorrect 6K & 0.182 & \uline{0.350} \\
\midrule
Hard 4K      & 0.137 & 0.296 \\
Easy 4K      & \uline{0.179} & \uline{0.352} \\
\bottomrule \\
\end{tabular}
\caption{
LCB scores of Qwen2.5 at two stages were further trained on four partitions of data. Easy data significantly outperforms hard data, but correctness has little impact.
}
\label{table:bar}
\end{minipage}

\end{table*}

\subsection{Datasets}
To answer RQ-1, we create three quantity-controlled datasets based on a single code reasoning dataset of 30,000 examples.
The questions are sourced from OpenCodeReasoning2 (OCR2)~\cite{ahmad2025opencodereasoning2}, which contains 34,125 unique competitive coding problems compiled from 4 data sources.
The answers are generated using two models, DeepSeek-R1-0528 and KAT-V1-40B~\cite{zhan2025kat}, with an average duplication count of 7, chosen because of their over $70\%$ accuracy on LCB.
As a reasoning dataset, each example contains a response that has \texttt{<think></think>} tags wrapped around the teacher model's thinking traces.

From this base set, we create a random subset of 10K examples, and then another random subset of 1K examples out of the 10K subset. 
We explicitly conduct random sampling to ensure the subsets follow the same distribution as the supersets and differ only in quantity.
We train a model of choice on the three datasets and report their LCB scores to study how dataset size impacts distillation outcomes.

To answer RQ-2, we take another in-distribution dataset of 13,583 reasoning examples whose questions are only sourced from the TACO~\cite{li2023taco} dataset. 
As TACO provides test cases for each question, we label each model response as either \texttt{correct} or \texttt{incorrect} based on whether it passes all test cases for the given question.
We then select two random subsets of 6K examples, one consisting of only correct responses and the other only incorrect responses.
The performance of models trained on these two separate datasets answers whether code correctness matters.

To answer RQ-3, we leverage the existing \texttt{difficulty} labels in TACO to partition examples into two groups:
examples labeled \texttt{hard}, \texttt{very\_hard}, or \texttt{medium\_hard} belong to the group of hard questions and the ones labeled \texttt{easy} or \texttt{medium} belong to the group of easy questions.
% Examples labeled \texttt{unknown\_difficulty} are discarded.
We randomly select from the two groups to create two mutually exclusive datasets of size 4K.
Models on the two datasets can provide useful signals on whether the model benefits from learning to solve harder questions. 

\subsection{Training setup}
We train on two small instruction-tuned models as the study of interest:
Qwen2.5-7B-Instruct (Qwen2.5) and Llama3.1-8B-Instruct (Llama3.1)~\cite{dubey2024llama}.
Neither has the ability to output CoT traces wrapped in \texttt{<think>} tags nor a dedicated special \texttt{<think>} token in the tokenizer.

For Qwen2.5, we conduct a total of eleven experiments, which consist of the three runs on the 1K, 10K, and 30K datasets to answer RQ-1, and four on the correctness- and difficulty-controlled datasets for both the instruct model and the 30K finetuned checkpoint (Qwen2.5-30K) to answer RQ-2 and RQ-3. 
Running the same four experiments on two checkpoints gives us signals on how the impact of the same data features changes as the model improves its reasoning capacity. 
For Llama3.1, we conduct three runs on the 1K, 10K, and 30K datasets to study the data scaling trend relevant to RQ-1.

We train each job using torchtune on $8\times$ Nvidia H100 GPUs. 
We uniformly use a global batch size of $128$, a learning rate of $8e-5$ with a warmup ratio of $0.10$, and the AdamW optimizer~\cite{loshchilov2017decoupled} to ensure fair comparisons.
We use a max sequence length of 32,768 due to Qwen's architecture constraints.
Each job takes a total of 5 epochs and is evaluated on the final checkpoint.
\section{Empirical Findings}
\label{sec:findings}

We now discuss the empirical findings that answer our three research questions in section \ref{sec:valley_of_reasoning}.

\textbf{RQ1: There is a "valley of code reasoning": the student model's distillation performance initially drops, then it increases as the training data quantity goes up.} 
As is illustrated in Fig \ref{fig:valley}, for both Qwen2.5 and Llama3.1, the same scaling trend holds:
The model's LCB score first decreases by more than half from the baseline when trained on 1K examples, then improves and surpasses the baseline by around $50\%$ when the data size reaches 10K.
On Qwen2.5 and Llama3.1, training on 30K examples further lifts the performance respectively by $50\%$ and $100\%$ upon the 10K checkpoint, exhibiting or even surpassing the anticipated log-linear trend of improvement.

We observe similar trends in two auxiliary metrics that correlate strongly with the quality of the model's reasoning outputs.
One of them is the \textit{completion rate}, defined as the percentage of all model responses that finish within 32K tokens.
The completion rate is highly correlated with effective reasoning as incomplete responses often repeat the same phrases before the eventual cutoff.
We observe a steady $1.65$-$1.9$ log-linear increase in the completion rate from the 1K to the 30K checkpoint.

Another metric is the \texttt{<think>} tag occurrence rate, defined as the percentage of all responses that begin with a \texttt{<think>} tag.
A higher \texttt{<think>} tag occurrence rate shows the model is effectively learning the structure of its outputs.
Surprisingly, the \texttt{<think>} tag is hard for the model to learn, as both models starts with below 20\%.
As data size scales up, we observe another log linear trend of $1.6$-$2.4$ increase in the \texttt{<think>} tag occurrence rate all the way to the later checkpoints.

\textbf{RQ2: Training examples that have correct responses do not improve small models' performances.}
For both Qwen2.5 and the SFT checkpoint with 30K data (Qwen2.5-30K), we look at the performance lift of further training on 6K correct responses versus 6K incorrect responses (Table~\ref{table:bar}).
Compared to each baseline, the two subsets both lift the LCB score by around $50\%$, which suggests that the correctness of responses has no effect on distillation outcomes at both stages of training. 

Furthermore, as the model continues to improve on the additional 6K at a rate comparable to the initial rise to 30K, it hints that model performance is far from saturation at the current 30K data scale.

% We also look at the perplexity loss of the models on the two sets of training data, which helps assess whether the model finds one of the two more difficult to learn (Fig 3).
% We find no statistically significant differences in the loss for the two datasets for both Qwen2.5 and Qwen2.5-30K.
% Together, the results suggest no factors distinguishing the dataset with all correct responses from the one with all incorrect responses from a training perspective.

\textbf{RQ3: Small models benefit more from easy and medium-level examples than from hard examples.}
For both Qwen2.5 and Qwen2.5-30K, we also compare their evaluation outcomes when trained on an additional 4K data with hard-coded questions versus 4K easy and medium ones (Table \ref{table:bar}).
For both models, training on exclusively hard questions provides only a modest boost:
Qwen2.5 improves by 7\%, whereas Qwen2.5-30K improves by 11\%.
In contrast, when training on only easy and medium-difficult data,
Qwen2.5 improves by 41\%, while Qwen2.5-30K improves by 33\%.
Therefore, we conclude that for small models in the medium-low data regime, examples containing easier questions provide superior benefits to downstream code reasoning performances.

% The inferior performance of models trained on harder data is also correlated with the fact that harder problems have a data distribution that is harder to learn. 
% Looking at the perplexity loss of models trained on difficult problems and easy problems respectively (Fig 3b), we observe that the loss starts at similar values before training takes part.
% At the end of training, the loss on the easier data reduces by around 48\% for Qwen2.5 and Qwen2.5-30K, while the loss on the harder data drops only by 30\%. 

Curiously, in all training runs that answer RQ-2 and RQ-3, we observe that the completion rate and the \texttt{<think>} tag occurrence rate are only weakly correlated with evaluation performances.
However, they are strongly correlated with data quantity, which jointly predicts the quality of code reasoning.
For example, for both Qwen2.5 and Qwen2.5-30K, training on hard questions yields the same completion rate and \texttt{<think>} tag occurrence rate as is in training on easy questions.
This suggests that the benefits of easier problems go beyond mere structural imitations of the teacher's reasoning traces.
\section{Conclusion}

In this work, we have demonstrated that for small LLMs in a low to medium-low data regime of around 1K-30K training samples, distillation performance does not scale monotonously with data quantity. 
Instead, the model performance follows a highly predictable pattern in the shape of a valley:
It initially decreases by half, then increases log-linearly with a higher completion rate and more \texttt{<think>} tag occurrences.
Backed by this observation, we study the impact of several data features in different stages of model training.
We show that for small LLMs across stages, reasoning examples with easier questions are more beneficial, but the correctness of model responses hardly matters.
We will take it upon ourselves to further work on showing how the trend scales up to the medium-high and high data regimes above 100K, and if the same conclusions about correctness and difficulty hold in those regions.

\bibliographystyle{collinear}
\bibliography{neurips_2025}

\end{document}